\documentclass[onecolumn,english,letterpaper, 10 pt, conference, twocolumn]{IEEEtran}
\usepackage[T1]{fontenc}
\usepackage[utf8]{inputenc}
\usepackage{xcolor}
\usepackage{colortbl}
\usepackage{array}
\usepackage{refstyle}
\usepackage{booktabs}
\usepackage{multirow}
\usepackage{amsmath}
\usepackage{amssymb}
\usepackage{graphicx}

\makeatletter

\newcommand*\LyXZeroWidthSpace{\hspace{0pt}}
\providecommand{\tabularnewline}{\\}
\RS@ifundefined{subsecref}{
  \newref{subsec}{
        name      = \RSsectxt,
        names     = \RSsecstxt,
        Name      = \RSSectxt,
        Names     = \RSSecstxt,
        refcmd    = {\S\ref{#1}},
        rngtxt    = \RSrngtxt,
        lsttwotxt = \RSlsttwotxt,
        lsttxt    = \RSlsttxt
  }
}{}

\usepackage{xcolor}         
\usepackage{pifont}         
\usepackage{cite}
\usepackage[caption=false]{subfig}
\usepackage{colortbl}
\usepackage{booktabs}
\usepackage{graphicx}
\usepackage{caption}

\definecolor{alpha}{HTML}{FFCE19}
\definecolor{bob}{HTML}{FF00FF}
\definecolor{carol}{HTML}{19FFDC}
\definecolor{Gray}{gray}{0.9} 

\usepackage[pagebackref,breaklinks,colorlinks]{hyperref}
\IEEEoverridecommandlockouts  

\ifdefined\showcaptionsetup
 \PassOptionsToPackage{caption=false}{subfig}
\fi
\usepackage{subfig}
\makeatother

\usepackage{babel}
\begin{document}
\title{GNIO: Gated Neural Inertial Odometry}
\author{Dapeng Feng$^{*}$, Yizhen Yin$^{*}$, Zhiqiang Chen, Yuhua Qi, and
Hongbo Chen\thanks{$^{*}$Equal contribution. Corresponding authors: Hongbo Chen.}\thanks{Dapeng Feng, Yizhen Yin, and Hongbo Chen are with Sun Yat-sen University,
Guangzhou, China. (e-mail: \{fengdp5, yinyzh5\}@mail2.sysu.edu.cn,
\{qiyh8, chenhongbo\}@mail.sysu.edu.cn)}\thanks{Zhiqiang Chen is with The University of Hong Kong, Hong Kong SAR,
China. (e-mail: zhqchen@connect.hku.hk)}}

\maketitle

\begin{abstract}
Inertial navigation using low-cost MEMS sensors is plagued by rapid
drift due to sensor noise and bias instability. While recent data-driven
approaches have made significant strides, they often struggle with
micro-drifts during stationarity and mode fusion during complex motion
transitions due to their reliance on fixed-window regression. In this
work, we introduce Gated Neural Inertial Odometry (GNIO), a novel
learning-based framework that explicitly models motion validity and
context. We propose two key architectural innovations: \ding{182}
a learnable Motion Bank that queries a global dictionary of motion
patterns to provide semantic context beyond the local receptive field,
and \ding{183} a Gated Prediction Head that decomposes displacement
into magnitude and direction. This gating mechanism acts as a soft,
differentiable Zero-Velocity Update (ZUPT), dynamically suppressing
sensor noise during stationary periods while scaling predictions during
dynamic motion. Extensive experiments across four public benchmarks
demonstrate that GNIO significantly reduces position drift compared
to state-of-the-art CNN and Transformer-based baselines. Notably,
GNIO achieves a $60.21\%$ reduction in trajectory error on the OxIOD
dataset and exhibits superior generalization in challenging scenarios
involving frequent stops and irregular motion speeds.
\end{abstract}

\section{Introduction}

Accurate state estimation and localization are fundamental prerequisites
for emerging technologies such as autonomous robotics, Virtual Reality
(VR), and Augmented Reality (AR). While Visual-Intertial Navigation
Systems (VINS) have become the de facto standard for high-precision
tracking, they face inherent limitations in challenging environments.
Visual methods are susceptible to failure in scenarios with extreme
lighting, motion blur, or visual occlusion, where conditions frequently
encountered in pedestrian navigation or pocket-constrained mobile
devices. To address these limitations, there has been a resurgence
of interest in Inertial Navigation Systems (INS) that rely solely
on strap-down Inertial Measurement Units (IMUs).

However, low-cost MEMS IMUs suffer from significant sensor noise and
bias instability. Traditional kinematic integration of these noisy
measurements leads to rapid drift, rendering pure physics-based dead
reckoning unusable within seconds.

To mitigate sensor errors without relying on external infrastructure
(such as GPS) or additional sensors (such as LiDAR), the research
community has pivoted towards data-driven approaches. Early deep learning
methods, such as IONet \cite{chen2018ionet} and RoNIN \cite{herath2020ronin},
demonstrated that neural networks could effectively learn to regress
2D velocity or displacement from windowed IMU data, significantly
outperforming traditional step-counting heuristics.

Building on this foundation, TLIO \cite{liu2020tlio} represented
a paradigm shift by moving from loosely coupled regression to a tightly
coupled filter framework. TLIO proposed a network that regresses the
estimated 3D displacement and, crucially, its associated uncertainties.
By fusing these learned relative state measurements into a Stochastic
Cloning Extended Kalman Filter (EKF), TLIO achieved robust 3D tracking
that combines the long-term stability of data-driven priors with the
short-term precision of kinematic models.

Since the introduction of TLIO, the field of Neural Inertial Odometry
(NIO) has advanced rapidly. Approaches such as AirIMU \cite{qiu2023airimu}
and AirIO \cite{qiu2025airio} have further refined uncertainty propagation
and feature extraction, enabling networks to learn more distinctive
motion representations. Recent works like EqNIO \cite{jayanth2024eqnio}
have introduced subequivariant architectures to better handle the
geometric symmetries of inertial data, ensuring that orientation changes
do not degrade estimation consistency. As highlighted by TinyIO \cite{zhang2025tinyio},
there is also a growing trend towards lightweight, reparameterized
models suitable for edge deployment. The recent IONext \cite{zhang2025ionext}
framework suggests that we are entering a new era where end-to-end
learning can capture increasingly complex motion dynamics.

Despite these advancements, a critical challenge remains in the adaptability
of the prediction mechanism itself. Most existing methods rely on
standard convolutional or recurrent backbones (such as ResNet \cite{he2016deep}
or GRU \cite{cho2014learning}) that regress displacement directly
from a fixed window of IMU data. These architectures often struggle
with two specific issues: \ding{182} Motion Ambiguity and Mode Confusion:
Human motion is highly variable, transitioning abruptly between stationary,
walking, and dynamic activities (e.g., running or climbing stairs).
A standard sliding window often lacks the memory to distinguish between
momentary jitter and the onset of a new motion mode, leading to erroneous
velocity updates. \ding{183} Oversensitive Regression: In scenarios
with low-magnitude motion or high sensor noise, standard regression
heads tend to output non-zero displacements due to the always-on nature
of the network. The existing methods attempt to mitigate this via
regressed uncertainty (covariance), but this is a reactive measure
rather than a proactive structural constraint. 

In this paper, we introduce GNIO (Gated Neural Inertial Odometry),
an enhanced learning-based framework built upon the robust tight-coupling
principles of TLIO \cite{liu2020tlio}. GNIO specifically addresses
the issues of motion mode confusion and oversensitive regression through
two architectural innovations:
\begin{enumerate}
\item Motion Bank: Unlike previous methods that treat IMU windows isolation
or with a limited recurrent state, GNIO incorporates a motion bank.
This mechanism maintains a compressed history of latent motion features,
allowing the network to attend to past motion patterns dynamically.
This enables the system to differentiate between consistent motion
trends and transient sensor artifacts, effectively providing a longer
temporal receptive field without the computational cost of processing
extremely long raw sequences.
\item Gated Prediction Head: We redesign the output layer of the neural
network to include a Gating Mechanism. Inspired by the gating units
in LSTMs and GRUs, this learnable gate acts as a value for the predicted
displacement. It explicitly decomposes the regression task into predicting
a magnitude scale (the gate) and a direction vector. When the system
detects stationarity or unreliable data, the gate actively suppresses
the displacement output towards zero, acting as a soft, learnable
ZUPT. During high-dynamic motion, the gate opens to allow full displacement
prediction.
\end{enumerate}
We demonstrate that integrating these mechanisms into the TLIO framework
results in superior position estimation accuracy and reduced drift,
particularly in complex trajectories involving frequent stops and
diverse motion speeds.

\section{Related Work}

Inertial Odometry (IO) has evolved from classical physical modeling
to data-driven approaches that leverage deep learning to correct sensor
bias and estimate motion. We categorize related works into pure learning-based
methods, hybrid filtering frameworks, and recent advancements in adaptive
mechanism design.

\subsection{Pure Data-Driven Inertial Odometry}

Early data-driven methods established the feasibility of regressing
motion directly from IMU measurements. IONet \cite{chen2018ionet}
addressed the curse of drift by segmenting inertial data into independent
windows and using Recurrent Neural Networks (RNNs) to estimate latent
states. This was followed by RoNIN \cite{herath2020ronin}, which
introduced a robust benchmark and explored ResNet \cite{he2016deep}
and TCN \cite{lea2017temporal} backbones for improved generalization.

Recent advancements have shifted toward capturing long-range dependencies
and fine-grained features. CTIN \cite{rao2022ctin} and iMoT \cite{nguyen2025imot}
leverage Transformer architectures with self-attention and progressive
decouplers to model complex temporal relationships. Further expanding
this, IONext \cite{zhang2025ionext} employs large-kernel convolutions
and adaptive dynamic mixers to balance global perception with local
sensitivity. 

\subsection{Hybird Filter-Integrated Frameworks}

A major trend in inertial navigation is the coupling of neural displacement
priors with classical state estimation filters. TLIO \cite{liu2020tlio}
pioneered this by integrating a deep network within a tightly-coupled
Extended Kalman Filter (EKF), using the network’s predicted displacements
as measurement updates to constrain the EKF's state.

Building on this, researchers have sought to incorporate physical
priors into the learning process. EqNIO \cite{jayanth2024eqnio} and
RIO \cite{cao2022rio} introduce rotation-equivariance and subequivariance,
ensuring that the network respects the physical symmetries of IMU
data (e.g., gravity-aligned rotation).

\subsection{Adaptive Motion Modeling and Gating Mechanisms}

RIDI \cite{yan2018ridi} initially identified that human motion consists
of discrete repetitive modes. More recently, StarIO \cite{zhang2025stario}
used high-dimensional nonlinear features and collaborative attention
to model complex motion dynamics, while WDSNet \cite{wang2024wavelet}
and FTIN \cite{zhang2025ftin} utilized wavelet selection and frequency-domain
integration to handle multi-scale signal variations.

GNIO occupies this space by introducing two critical modifications
to the TLIO \cite{liu2020tlio} prediction head:
\begin{enumerate}
\item Motion bank: Drawing inspiration from the pattern-recognition capabilities
of iMoT \cite{nguyen2025imot} and StarIO \cite{zhang2025stario},
the motion bank in GNIO specific motion modes from historical features.
This enables the model to apply specialized weights tailored to different
movement types, effectively curing drift through adaptive prediction.
\item Gated Mechanism: Unlike standard regression heads that output values
directly, GNIO employs a gating structure to explicitly modulate the
magnitude and direction of the predicted displacement. This allows
the network to suppress noise or amplify the signal based on the context
of the motion.
\end{enumerate}

\section{Methods}

\subsection{Overview}

\begin{figure*}
\begin{centering}
\includegraphics[width=1\linewidth]{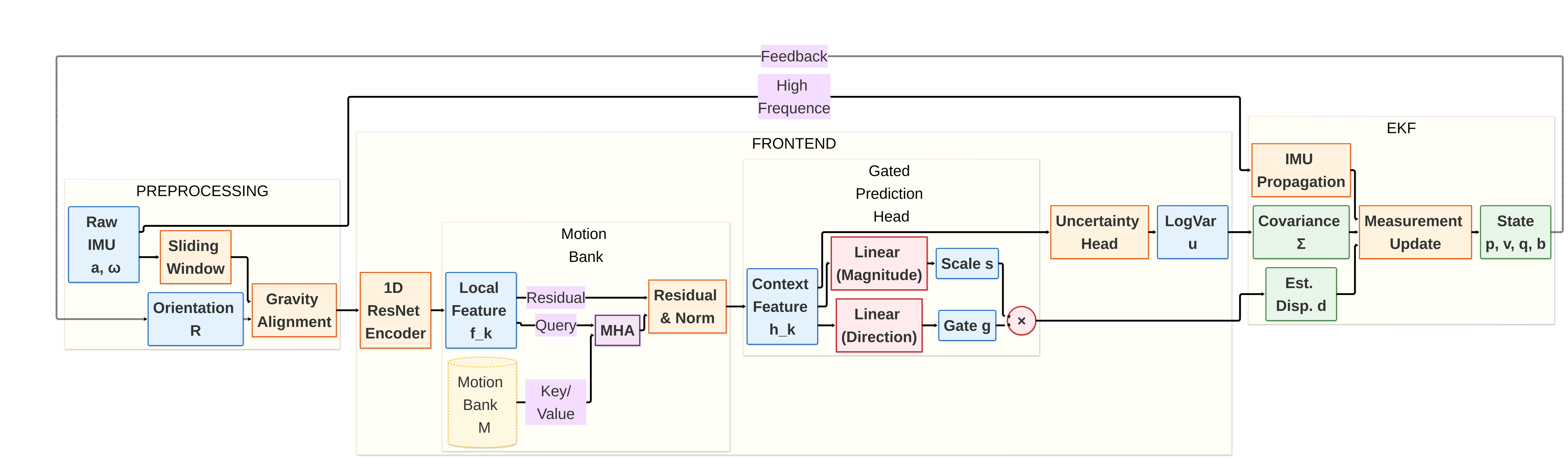}
\par\end{centering}
\caption{\label{fig:System-Overview-of}System Overview of GNIO.}
\end{figure*}

Our proposed GNIO framework builds upon the tightly-coupled archiecture
of TLIO \cite{liu2020tlio}. As shown in \Figref{System-Overview-of},
the system consists of two main components: a Neural Front-End and
an Extened Kalman Filter (EKF).

The Neural Front-End receives a window of gravity-aligned IMU data
and regresses a relative 3D displacement measurement $\hat{\mathbf{d}}_{k}$
along with its associated uncertainty covariance $\hat{\Sigma}_{k}$.
These learned measurements are used to update the EKF state.

Unlike TLIO, which uses a standard ResNet \cite{he2016deep}, GNIO
introduces a Motion Bank and a Gated Prediction Head to dyanmically
adapt to complex motion modes and explicitly model stationarity.

\subsection{Network Input and Feature Extraction}

Let the raw IMU measurements at time stemp $t$ be denoted as $\mathbf{m}_{t}=\left[\mathbf{a}_{t},\boldsymbol{\mathbf{\omega}}_{t}\right]\in\mathbf{\mathbb{R}}^{6}$.
Following TLIO, we extract a sliding window of N samples. To ensure
the network learns motion physics independent of global yaw, we rotate
the inputs into a gravity-aligned frame using the current EKF orientation
estimate $\mathbf{R}^{wb}_{k}$:
\begin{equation}
\mathbf{x}_{i}=\mathbf{R}^{wb}_{k}\left(\mathbf{m}_{i}-\mathbf{b}_{k}\right),\quad i\in\left[t-N,t\right].
\end{equation}

The input tensor $\mathbf{X}_{k}\in\mathbb{R}^{N\times6}$ is fed
into a 1D ResNet backbone (similar to ResNet-18) to extract a high-dimensional
feature $\mathbf{f}_{k}\in\mathbb{R}^{D}$:

\begin{equation}
\mathbf{f}_{k}=\text{Encoder}\left(\mathbf{X}_{k}\right).
\end{equation}

\subsection{Motion Bank}

Standard sliding-window approaches often suffer from mode confusion
(e.g., distinguishing between standing still and slow walking) due
to limited receptive fields. To address this, we introduce a Motion
Bank containing global motion patterns.

Unlike recurrent architectures that rely on a transient hidden state,
our motion bank is a learnable parameter matrix $\mathcal{M}\in\mathbb{R}^{m\times D}$,
where $m$ represents the number of global motion prototypes (\textit{e.g.},
static, walking, running, turning) and D is the feature dimension.
These patterns are learned end-to-end during training and serve as
a static knowledge base.

We employ a Multi-Head Attention (MHA) \cite{vaswani2017attention}
mechanism to contextually query this global knowledge base. The current
local feature $\mathbf{f}_{k}$, extracted by the ResNet encoder,
acts as the Query($\mathbf{Q}$), representing the specific kinematic
state at the current time step. The global motion patterns in $\mathcal{M}$
serve as both the Keys($\mathbf{K}$) and Values($\mathbf{V}$). Formally,
the query, key, and value projections are computed as:
\begin{align}
\mathbf{Q} & =\mathbf{f}_{k}\mathbf{W}_{Q},\\
\mathbf{K} & =\mathbf{M}\mathbf{W}_{K},\\
\mathbf{V} & =\mathbf{M}\mathbf{W}_{V},
\end{align}
where $\mathbf{W}_{Q},\mathbf{W}_{K},\mathbf{W}_{V}$ are learnable
projection matrices. This formulation allows the network to compare
the transient local features against the stable global prototypes.

The relevance of each global motion pattern to the current input is
determined by the attention weights, calculated via the scaled dot-product.
The context $\mathbf{c}_{k}$ is then derived as a weighted sum of
the prototypes, effectively reconstructing a denoised or idealized
representation of the current motion:
\begin{equation}
\mathbf{c}_{k}=\text{Softmax}\left(\frac{\left(\mathbf{f}_{k}\mathbf{W}_{Q}\right)\left(\mathbf{M}\mathbf{W}_{K}\right)^{T}}{\sqrt{d_{k}}}\right)\left(\mathbf{M}\mathbf{W}_{V}\right).
\end{equation}

Intuitively, if the input features $\mathbf{f}_{k}$ strongly resemble
a stationary prototype stored in $\mathbf{M}$, the attention mechanism
will retrieve a context $\mathbf{c}_{k}$ that emphasizes zero-motion
characteristics, guiding the downstream components to suppress drift.

Finally, the retrieved global context is fused with the original local
features to produce the enriched representation $\mathbf{h}_{k}$.
We employ a residual connection to preserve high-frequency details
from the local window while incorporating the semantic guidance from
the global context:
\begin{equation}
\mathbf{h}_{k}=\mathbf{f}_{k}+\mathbf{c}_{k}.
\end{equation}

This fused feature $\mathbf{h}_{k}$ serves as the input to the subsequent
Gated Prediction Head, ensuring that displacement predictions are
informed by both immediate sensor readings and long-term structural
motion priors.

\subsection{Gated Prediction Head}

Standard regression heads in NIO models operate in an always-on manner,
continuously mapping features to displacement estimates. This architecture
often struggles to output true zero values during stationarity, leading
to accumulated drift from sensor noise, or micro-movements. Furthermore,
a single linear mapping may lack the flexibility to adjust the scale
of predictions dynamically across diverse motion intensities. To magnitude
these issues, we propose a Gated Prediction Head that explicitly decomposes
the regression task into magnitude modulation and directional estimation.

We design the motion prediction head as a dual-branch structure operating
on the context-enriched feature $\mathbf{h}_{k}$. The first branch
focuses solely on the magnitude of displacement. We apply a linear
transformation followed by a Softplus activation function to enforce
non-negativity, interpreting the output $\tilde{\mathbf{s}}_{k}\in\mathbb{R}^{3}$
as a scale or speed proposal for each axis:
\begin{equation}
\tilde{\mathbf{s}}_{k}=\text{Softplus}\left(\mathbf{W}_{s}\mathbf{h}_{k}+\mathbf{b}_{s}\right).
\end{equation}
Simultaneously, the second branch computes a directional gate $\mathbf{g}_{k}\in\mathbb{R}^{3}$.
We employ the Tanh activation function for this branch, constraining
its output to the range $\left[-1,1\right]$:
\begin{equation}
\mathbf{g}_{k}=\text{Tanh}\left(\mathbf{W}_{g}\mathbf{h}_{k}+\mathbf{b}_{g}\right).
\end{equation}
This gate serves a dual purpose: its sign determines the direction
of movement along each axis, while its magnitude acts as a confidence
score or attenuation factor. During stationary periods, the network
learns to drive $\mathbf{g}_{k}\rightarrow0$, effectvely suppressing
any noise from the magnitude branch. The final predicted 3D displacement
$\hat{\mathbf{d}}_{k}\in\mathcal{\mathbb{R}}^{3}$ is computed as
the element-wise product:
\begin{equation}
\hat{\mathbf{d}}_{k}=\tilde{s}_{k}\odot\mathbf{g}_{k}.
\end{equation}

\subsection{Uncertainty Estimation}

Consistent with the TLIO \cite{liu2020tlio}, we must also estimate
the reliability of our prediction for the EKF update. We regress the
log-variance $\mathbf{u}_{k}\in\mathbb{R}^{3}$ from the fused features
$\mathbf{h}_{k}$:
\begin{equation}
\mathbf{u}_{k}=\mathbf{W}_{u}\mathbf{h}_{k}+\mathbf{b}_{u}.
\end{equation}
The convariance matrix $\hat{\Sigma}_{k}$ is constructed assuming
a diagonal distribution:
\begin{equation}
\hat{\Sigma}_{k}=\text{diag}\left(\exp\left(2\mathbf{u}_{k,x}\right),\exp\left(2\mathbf{u}_{k,y}\right),\exp\left(2\mathbf{u}_{k,z}\right)\right).
\end{equation}
This formulation ensures numerical stability and positivity of the
variance.

\subsection{Loss Function}

To enable stable convergence of the displacement regression while
learning the heteroscedastic uncertainty required for the EKF update,
we employ a two-stage training strategy using Mean Squared Error (MSE)
and Negative Log-Likelihood (NLL).

The total loss $\mathcal{L}_{\text{total}}$ is a weighted sum of
the displacement error term and the probabilistic uncertainty term:
\begin{equation}
\mathcal{L}_{\text{total}}=\lambda_{\text{MSE}}\mathcal{L}_{\text{MSE}}+\lambda_{\text{NLL}}\mathcal{L}_{\text{NLL}},
\end{equation}

\noindent where $\mathcal{L}_{\text{MSE}}=\lVert\mathbf{d}^{gt}_{k}-\hat{\mathbf{d}}_{k}\rVert^{2}$
stabilizes the initial convergence of the displacement branches (Magnitude
and Gate), preventing large gradients from the uncertainty estimation
early in training. The Gaussian NLL term $\mathcal{L}_{\text{NLL}}$
ensures the predicted convariance $\hat{\Sigma}_{k}$ aligns with
the actual residual error distribution:
\begin{equation}
\mathcal{L}_{\text{NLL}}=\frac{1}{2}\lVert\mathbf{d}^{gt}_{k}-\hat{\mathbf{d}}_{k}\rVert^{2}_{\hat{\Sigma}}+\frac{1}{2}\log\det\left(\hat{\Sigma}_{k}\right).
\end{equation}

This simultaneous optimization encourages the network to minimize
displacement error while accurately quantifying its confidence, which
is critical for the tight fusion mechanism in the EKF.

\subsection{Stochastic Cloning EKF}

We adopt the Stochastic Cloning EKF from TLIO \cite{liu2020tlio}
to tightly fuse the learned relative displacements with high-frequency
IMU kinematics.

\subsubsection{State Representation}

The filter tracks the IMU state (position, velocity, orientation,
and sensor biases) along with a sliding window of cloned past poses.

\subsubsection{Measurement Update}

When the Neural Front-End predicts a 3D displacement $\hat{\mathbf{d}}_{ij}$
and its covariance $\hat{\Sigma}_{ij}$ \LyXZeroWidthSpace , we apply
a measurement update to the filter. The residual $\mathbf{r}$ is
defined as:
\begin{equation}
\mathbf{r}=\hat{\mathbf{d}}_{ij}-\mathbf{R}^{T}_{\gamma i}\left(\mathbf{p}_{j}-\mathbf{p}_{i}\right),
\end{equation}
 where $\mathbf{p}_{i},\mathbf{p}_{j}$ are the cloned positions and
$\mathbf{R}_{\gamma i}$ represents the gravity-aligned yaw rotation.

\section{Experiments}

\subsection{Implementation Details}

The GNIO model is trained with Adam \cite{kingma2014adam} optimizer
with $\beta_{1}=0.9,\beta_{2}=0.999$. The batch size is set to $1024$.

\subsubsection{Training Strategy}

To stabilize the early stages of training and prevent large gradients
from the attention mechanism, we employ a Linear Warm-Up phase followed
by a Cosine Annealing learning rate schedule. The initial learning
rate starts at $1\times10^{-6}$ and linearly increases to $1\times10^{-4}$
over the first $5$ epochs. Subsequently, it decays to a minimum of
$1\times10^{-6}$ over the remaining training duration, which is $200$
epochs in total. We employ a composite loss function $\mathcal{L}_{\text{total}}$
to balance displacement accuaracy and uncertainty consistency. We
empirically set the weighting factors to $\lambda_{\text{MSE}}=1\times10^{2}$
and $\lambda_{\text{NLL}}=1\times10^{-4}$.

\subsubsection{Data Preprocessing}

Input IMU sequences are processed using a sliding window approach
designed to maintain a consistent temporal receptive field and overlap
across varying sensor hardware. We fix the temporal duration of each
window to 1.0 second and the stride between consecutive windows to
0.1 seconds. Consequently, the window size $N$ and stride $S$ are
dynamically adjusted based on the IMU sampling frequency:
\begin{itemize}
\item For $100\text{ Hz}$ IMUs: $N=100$ samples, $S=10$ samples.
\item For $200\text{ Hz}$ IMUs: $N=200$ samples, $S=20$ samples.
\end{itemize}
This consistency ensures that the Motion Bank and Gated Prediction
Head receive a standardized duration of kinematic information to identify
motion modes. 

\subsubsection{Network Configuration}

The Motion bank size is set to $m=64$ (the number of learnable global
motion prototypes), which we found provides an optimal balance between
expressivity and redundancy (see \Subsecref{Ablation-Study}). The
dimension of the high-dimensional latent feature space is $D=512$.
The backbone follows a 1D ResNet-18 architecture modified for inertial
signal processing.

\subsection{Datasets}

To evaluate the generalization capability of GNIO across diverse sensor
placements and motion behaviors, we utilize five public datasets established
in recent literature \cite{herath2020ronin,liu2020tlio,yan2018ridi,chen2018oxiod,sun2021idol}.
These datasets cover a wide spectrum of human activities, ranging
from steady walking to complex irregular movements.

\subsubsection{OxIOD \cite{chen2018oxiod}}

A comprehensive dataset capturing everyday activities. The dynamics
encompass a mix of slow walking, normal walking, running, and even
trolley pushing, providing variations in speed and periodicity crucial
for testing the Gated Prediction Head.

\subsubsection{RIDI \cite{yan2018ridi}}

Features smartphone data captured in a Vicon-equipped room. It emphasizes
placement diversity (body-attached, bag, hand) and includes complex,
erratic movements with frequent directional changes, designed to test
robustness against placement shifts.

\subsubsection{RoNIN \cite{herath2020ronin}}

Collected with Android smartphones across three carrying modes (handheld,
pocket, bag). The motion dynamics include natural, unconstrained walking
and running by $100$ distinct subjects, providing a large-scale benchmark
for inter-subject generalization.

\subsubsection{IDOL \cite{sun2021idol}}

Focuses on challenging orientation estimation scenarios. The motion
dynamics involve sharp turns, sudden stops, and burst movements, which
test the system's ability to maintain heading consistency during rapid
maneuvers.

\subsubsection{TLIO \cite{liu2020tlio}}

Acquired using a VR headset in indoor environments. The dynamics are
characterized by smooth head motions, coupled with complex 3D trajectories
involving stair climbing and multi-floor navigation, challenging the
vertical tracking capabilities.

\subsection{Baselines}

To rigorously evaluate the performance of GNIO, we compare it against
four state-of-the-art learning-based inertial odometry methods, ranging
from foundational CNNs to advanced Transformers and lightweight edge-optimized
models. 

\begin{table}
\caption{\label{tab:Comparison-of-model}Comparison of model parameters and
computational complexity.}

\resizebox{\linewidth}{!}{
\begin{centering}
\begin{tabular}{llcc}
\toprule 
\textbf{Architecture} & \textbf{Model Family} & \textbf{Params (M)} & \textbf{GFLOPs}\tabularnewline
\midrule 
\rowcolor{lightgray}\multicolumn{4}{l}{\textit{CNN-based}}\tabularnewline
TLIO & ResNet-18+EKF & $5.42$ & $0.08$\tabularnewline
DeepILS & Lightweight ResNet & $2.29$ & $0.03$\tabularnewline
\textbf{GNIO (Ours)} & Memory + Gated & \textbf{$\boldsymbol{4.90}$} & \textbf{$\boldsymbol{0.08}$}\tabularnewline
\midrule 
\rowcolor{lightgray}\multicolumn{4}{l}{\textit{Transformer-based}}\tabularnewline
CTIN & Contextual Transformer & $0.56$ & $7.27$\tabularnewline
iMoT & Seq-to-Seq Transformer & $14.49$ & $7.79$\tabularnewline
\bottomrule
\end{tabular}
\par\end{centering}
}
\end{table}

\subsubsection{TLIO \cite{liu2020tlio}}

The foundational tightly-coupled framework that uses a ResNet-based
regressor within an EKF. As the direct predecessor to GNIO, it serves
as the primary baseline to demonstrate the impact of our architectural
innovations.

\subsubsection{CTIN \cite{rao2022ctin}}

A contextual Transformer network that replaces standard convolutions
with a transformer architecture. It leverages self-attention to capture
long-range temporal dependencies, providing a strong baseline for
context-aware modeling.

\subsubsection{iMoT \cite{nguyen2025imot}}

The intertial motion transformer is a recent approach using a full
encoder-decoder transformer structure. It represents the current state-of-the-art
in sequence-to-sequence learning for inertial navigation, offering
a rigorous benchmark for complex motion dynamics.

\subsubsection{DeepILS \cite{tariq2025deepils}}

A recently proposed lightweight framework designed for efficient edge
deployment. DeepILS utilizes a residual network enhanced with channel-wise
and spatial attention mechanisms to achieve high accuracy with low
computational cost.

To contextualize the performance of GNIO within the constraints of
real-world robotic and mobile deployment, we provide a quantitative
comparison of the computational complexity for all evaluated models
in \Tabref{Comparison-of-model}. By benchmarking against this diverse
set of models, we demonstrate that GNIO achieves a superior balance
between the lightweight efficiency of traditional CNNs and the high-accuracy
contextual reasoning typically reserved for much heavier Transformer
architectures.

\begin{table*}
\caption{\label{tab:Trajectory-Prediction-Comparisio}Trajectory Prediction
Comparision of GNIO and Baseline Models.}

\resizebox{\linewidth}{!}{
\begin{centering}
\begin{tabular}{l>{\columncolor{Gray}\centering}c>{\columncolor{Gray}\centering}ccc>{\columncolor{Gray}\centering}c>{\columncolor{Gray}\centering}ccc}
\toprule 
\multirow{2}{*}{Methods} & \multicolumn{2}{>{\columncolor{Gray}\centering}c}{OxIOD \cite{chen2018oxiod}} & \multicolumn{2}{c}{RIDI \cite{yan2018ridi}} & \multicolumn{2}{>{\columncolor{Gray}\centering}c}{RoNIN \cite{herath2020ronin}} & \multicolumn{2}{c}{IDOL \cite{sun2021idol}}\tabularnewline
 & seen & unseen & seen & unseen & seen & unseen & seen & unseen\tabularnewline
\midrule
TLIO \cite{liu2020tlio} & $2.67$ & $2.54$ & $1.58$ & $1.49$ & $5.95$ & $7.10$ & $5.$89 & $5.34$\tabularnewline
CTIN \cite{rao2022ctin} & $2.32$ & $3.34$ & $1.39$ & $1.86$ & $4.62$ & $5.61$ & $2.90$ & $3.69$\tabularnewline
iMoT \cite{nguyen2025imot} & $1.86$ & $0.90$ & $1.68$ & $1.49$ & $\boldsymbol{3.78}$ & $5.31$ & \textbf{$\boldsymbol{2.22}$} & $\boldsymbol{3.00}$\tabularnewline
DeepILS \cite{tariq2025deepils} & -- & -- & $1.56$ & $1.48$ & $4.80$ & $6.80$ & -- & --\tabularnewline
\midrule
\textbf{GNIO (Ours)} & $\boldsymbol{0.74}^{\uparrow60.21\%}$ & $\boldsymbol{0.71}^{\uparrow21.11\%}$ & $\boldsymbol{1.22}^{\uparrow12.23\%}$ & $\boldsymbol{1.19}^{\uparrow19.59\%}$ & $4.31^{\downarrow14.02\%}$ & $\boldsymbol{5.02}^{\uparrow5.46\%}$ & $3.45^{\downarrow55.45\%}$ & $3.89^{\downarrow29.66\%}$\tabularnewline
\bottomrule
\end{tabular}
\par\end{centering}
}
\end{table*}

\subsection{Comparative Results and Analysis}

We evaluate GNIO against four representative baselines: the foundational
TLIO \cite{liu2020tlio}, the contextual transformer CTIN \cite{rao2022ctin},
the full motion transformer iMoT \cite{nguyen2025imot}, and the lightweight
DeepILS \cite{tariq2025deepils}. \Tabref{Trajectory-Prediction-Comparisio}
summarizes the Absolute Trajectory Error (ATE) in meters across four
major benchmarks. \textbf{Bold} values indicate the overall state-of-the-art
(SOTA) performance in each category. The results demonstrate GNIO's
superior generalization and its ability to significantly cure the
drift inherent in traditional regression models.

\subsubsection{Dominance in Pedestrian Benchmarks}

GNIO shows its most significant improvements on the OxIOD \cite{chen2018oxiod}
and RIDI \cite{yan2018ridi} datasets, which are characterized by
diverse pedestrian activities and varied phone placements.
\begin{itemize}
\item \textbf{OxIOD}: GNIO achieves an ATE of $0.74m$, surpassing the runner-up
(iMoT, $1.86m$) by a massive margin of $60.21\%$. This demonstrates
that when the environment and motion types are known, the Gated Prediction
Head's ability to decompose magnitude and direction allows for near-perfect
soft-ZUPT execution.
\item \textbf{RIDI}: In the more challenging unseen split, GNIO leads the
runner-up (DeepILS, $1.48m$) by $19.59\%$. This validates that the
Motion Bank effectively generalizes a dictionary of human motion that
remains valid across different subjects and sensor placements.
\end{itemize}

\subsubsection{Robustness vs. Model Capacity}

The RoNIN \cite{herath2020ronin} dataset reveals a critical trade-off
between raw model capacity and generalization.
\begin{itemize}
\item \textbf{Seen Split}: The complex encoder-decoder structure of iMoT
maintains the state-of-the-art at $3.78m$, with GNIO trailing by
$14.02\%$.
\item \textbf{Unseen Split}: Crucially, when transitioning to the unseen
split, GNIO takes the lead, outperforming iMoT ($5.31m$) by $5.46\%$.
\end{itemize}
This reversal suggests that while deep transformers like iMoT can
overfit to the specific motion signatures of the training subjects,
GNIO’s use of a ResNet-18 backbone coupled with a Motion Bank provides
a more stable prior that generalizes better to the wild movements
of 100+ distinct subjects.

\subsubsection{Performance Gaps in High-Dynamic Scenarios}

On the IDOL \cite{sun2021idol} dataset, which is characterized by
sharp turns and rapid orientation changes, GNIO significantly improves
upon its direct predecessor TLIO ($5.34m$ unseen to $3.89m$ unseen).
However, it currently trails the state-of-the-art iMoT by $29.66\%$
in the unseen split.

While GNIO's gating mechanism excels at suppressing micro-drift during
stationarity, the results in IDOL suggest that full-sequence attention
(as seen in iMoT) remains more effective for capturing the aggressive
dynamics of burst movements. Nevertheless, GNIO offers a more computationally
efficient middle ground by augmenting a standard ResNet backbone with
its targeted architectural innovations.

\subsection{\label{subsec:Ablation-Study}Ablation Study}

To investigate the individual contributions of the proposed architectural
components, we conducted a systematic ablation study on the TLIO dataset.
We incrementally integrated the Motion Bank and the Gated Prediction
Head into the baseline TLIO architecture to evaluate their impact
on the Root Mean Square Error (RMSE) of the estimated trajectory.

\subsubsection{Impact of Key Components}

\Tabref{Component-wise-Ablation-Analysis} presents the performance
comparison of different model configurations.
\begin{itemize}
\item Baseline (TLIO): The standard ResNet-based regressor achieves an RMSE
of $2.40m$.
\item Effect of Motion Bank: Introducing the Motion Bank reduces the RMSE
of $1.98m$. This demonstrates that querying global motion prototypes
helps the network resolve ambiguities in local sensor data, providing
better semantic context for the regression.
\item Effect of Gated Prediction Head: Integrating only the Gated Prediction
Head yields an RMSE of $1.84m$, a significant improvement over the
baseline. This confirms that the decoupling of magnitude and direction
allows for more precise control over displacement, particularly in
suppressing drift during stationarity.
\item Full GNIO Framework: Combining both components leads to the best performance
with an RMSE of $1.66m$, validating that the Motion Bank and Gated
Head are complementary: one provides the necessary context, while
the other enforces structural consistency.
\end{itemize}
\begin{table}
\caption{\label{tab:Component-wise-Ablation-Analysis}Component-wise Ablation
Analysis on TLIO Dataset.}

\resizebox{\linewidth}{!}{
\begin{centering}
\begin{tabular}{lcccc}
\toprule 
\textbf{Configuration} & \textbf{Motion Bank} & \textbf{Gated Prediction Head} & \textbf{RMSE ($\boldsymbol{m}$)} & \textbf{Gain ($\boldsymbol{\%}$)}\tabularnewline
\midrule
Baseline (TLIO) & -- & -- & $2.40$ & --\tabularnewline
Context Enriched & $\checkmark$ & -- & $1.98$ & $17.5$\tabularnewline
Structurally Gated & -- & $\checkmark$ & $1.84$ & $23.3$\tabularnewline
\textbf{GNIO (Full)} & $\checkmark$ & $\checkmark$ & $\boldsymbol{1.66}$ & $\boldsymbol{30.8}$\tabularnewline
\bottomrule
\end{tabular}
\par\end{centering}
}
\end{table}

\subsubsection{Analysis of Gating Activation Functions}

We further investigated the design choices within the Gated Prediction
Head, specifically the activation function for the gate ($\sigma$)
and the function used for the magnitude scale. As shown in \Tabref{Evaluation-of-Gating},
we compared the standard Sigmoid activation against Tanh, and different
scale functions.
\begin{itemize}
\item \textbf{Sigmoid vs. Tanh:} Using Tanh ($\mathbf{g}_{k}\in\left[-1,1\right]$)
outperforms Sigmoid ($\mathbf{g}_{k}\in\left[0,1\right]$), reducing
RMSE from $2.05m$ to $2.00m$. This suggests that the ability to
invert the direction of the raw proposal is crucial for correcting
sign errors in the feature extraction.
\item \textbf{Scale Function:} While the absolute value (abs) and exponential
(exp) functions provided significant gains over the baseline, we found
that smooth, non-linear mappings yielded the best results. Specifically,
the \textbf{Softplus} and \textbf{Positive ELU} functions achieved
the lowest errors of $1.84m$ and $1.85m$, respectively. These functions
provide a more stable gradient flow compared to the non-differentiable
abs at zero or the extreme sensitivity of the exp function.
\end{itemize}
\begin{table}
\caption{\label{tab:Evaluation-of-Gating}Design Ablation of the Gated Prediction
Head on the TLIO Dataset.}

\resizebox{\linewidth}{!}{
\begin{centering}
\begin{tabular}{llcc}
\toprule 
\textbf{Gating $\boldsymbol{\sigma\left(\cdot\right)}$} & \textbf{Scaling $\boldsymbol{\psi\left(\cdot\right)}$} & \textbf{Scale Formula} & \textbf{RMSE ($\boldsymbol{m}$)}\tabularnewline
\midrule 
\rowcolor{lightgray}\multicolumn{4}{l}{\textit{Gating Activation Analysis}}\tabularnewline
Sigmoid & Linear & $x$ & $2.05$\tabularnewline
Tanh & Linear & $x$ & $2.00$\tabularnewline
\midrule 
\rowcolor{lightgray}\multicolumn{4}{l}{\textit{Magnitude Mapping Analysis (Fixed Tanh Gating)}}\tabularnewline
Tanh & Exp & $e^{x}$ & $1.94$\tabularnewline
Tanh & Abs & $\lvert x\rvert$ & $1.92$\tabularnewline
Tanh & PosELU & $\text{ELU}\left(x\right)+1$ & $1.85$\tabularnewline
\textbf{Tanh} & \textbf{Softplus} & $\ln\left(e^{x}+1\right)$ & $\boldsymbol{1.84}$\tabularnewline
\bottomrule
\end{tabular}
\par\end{centering}
}
\end{table}

\begin{figure}
\begin{centering}
\includegraphics[width=0.7\linewidth]{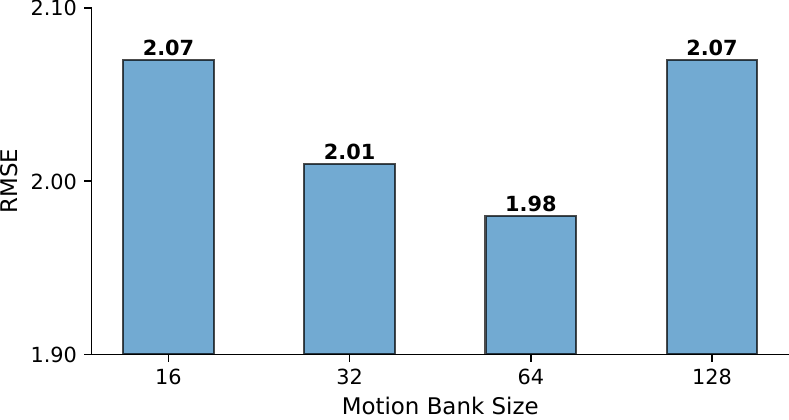}
\par\end{centering}
\caption{\label{fig:Impact-of-Motion}Impact of Motion Bank Size on Trajectory
Error}
\end{figure}

\begin{figure*}
\begin{minipage}[t]{0.23\linewidth}%
\subfloat[\label{fig:OxIOD-SLOW_WALKING}OxIOD-SLOW\_WALKING]{\begin{centering}
\includegraphics[width=1\linewidth,totalheight=5cm,keepaspectratio]{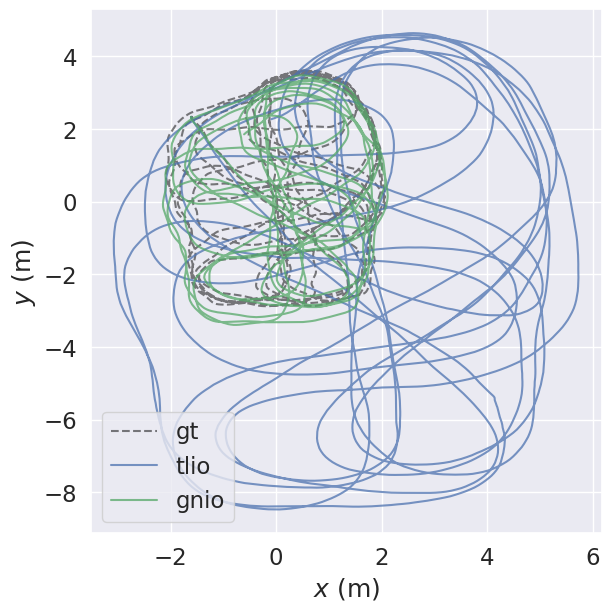}
\par\end{centering}
}%
\end{minipage}\hfill{}%
\begin{minipage}[t]{0.23\linewidth}%
\subfloat[\label{fig:OxIOD-HANDBAG}OxIOD-HANDBAG]{\begin{centering}
\includegraphics[width=1\linewidth,totalheight=5cm,keepaspectratio]{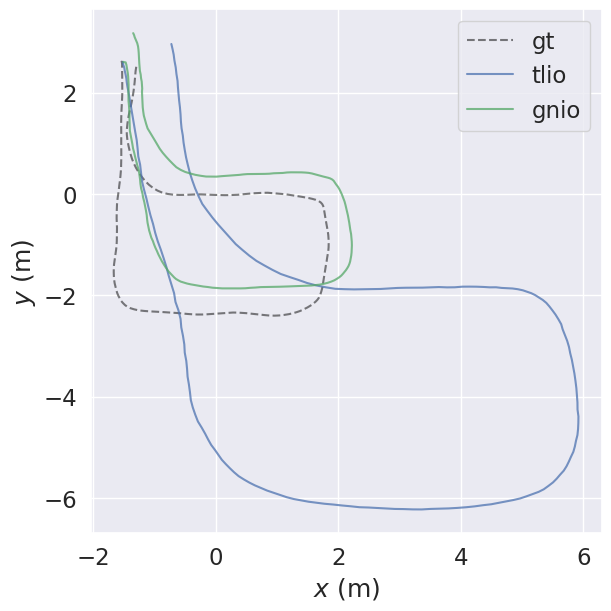}
\par\end{centering}
}%
\end{minipage}\hfill{}%
\begin{minipage}[t]{0.23\linewidth}%
\subfloat[\label{fig:RIDI-ZHICHENG_HANDHELD2}RIDI-ZHICHENG\_HANDHELD2]{\begin{centering}
\includegraphics[width=1\linewidth,totalheight=5cm,keepaspectratio]{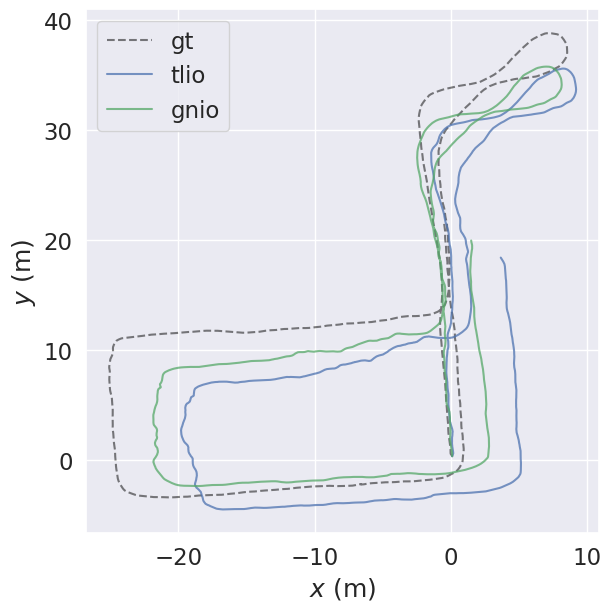}
\par\end{centering}
}%
\end{minipage}\hfill{}%
\begin{minipage}[t]{0.23\linewidth}%
\subfloat[\label{fig:RIDI-SHALI_BAG1}RIDI-SHALI\_BAG1]{\begin{centering}
\includegraphics[width=1\linewidth,totalheight=5cm,keepaspectratio]{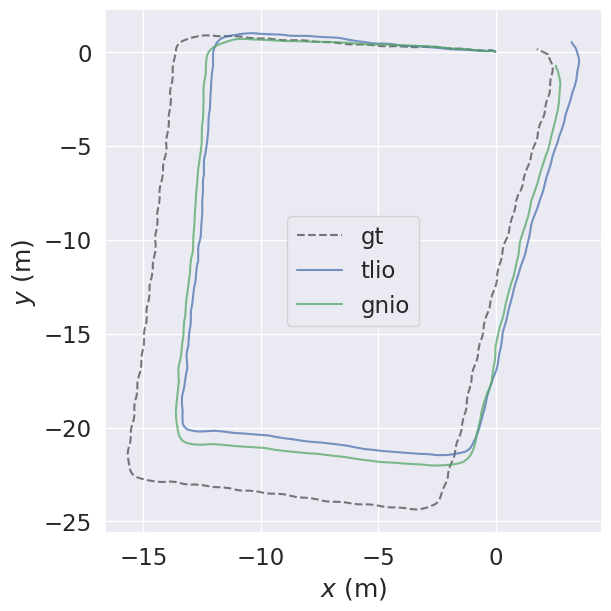}
\par\end{centering}
}%
\end{minipage}\\
\begin{minipage}[t]{0.23\linewidth}%
\subfloat[\label{fig:RoNIN-A016_1}RoNIN-A016\_1]{\begin{centering}
\includegraphics[width=1\linewidth,totalheight=5cm,keepaspectratio]{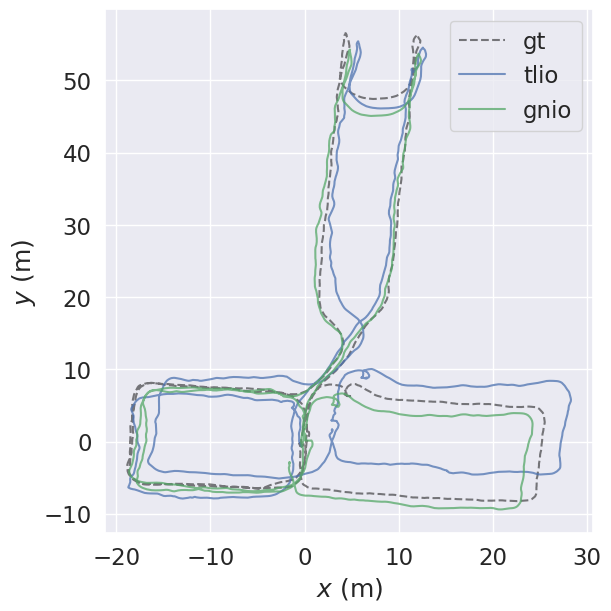}
\par\end{centering}
}%
\end{minipage}\hfill{}%
\begin{minipage}[t]{0.23\linewidth}%
\subfloat[\label{fig:RoNIN-A052_2}RoNIN-A052\_2]{\begin{centering}
\includegraphics[width=1\linewidth,totalheight=5cm,keepaspectratio]{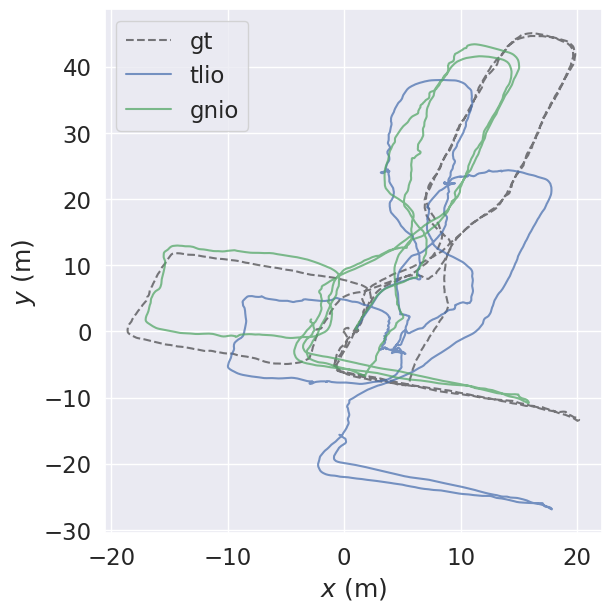}
\par\end{centering}
}%
\end{minipage}\hfill{}%
\begin{minipage}[t]{0.23\linewidth}%
\subfloat[\label{fig:IDOL-B1_KNOWN_8}IDOL-B1\_KNOWN\_8]{\begin{centering}
\includegraphics[width=1\linewidth,totalheight=5cm,keepaspectratio]{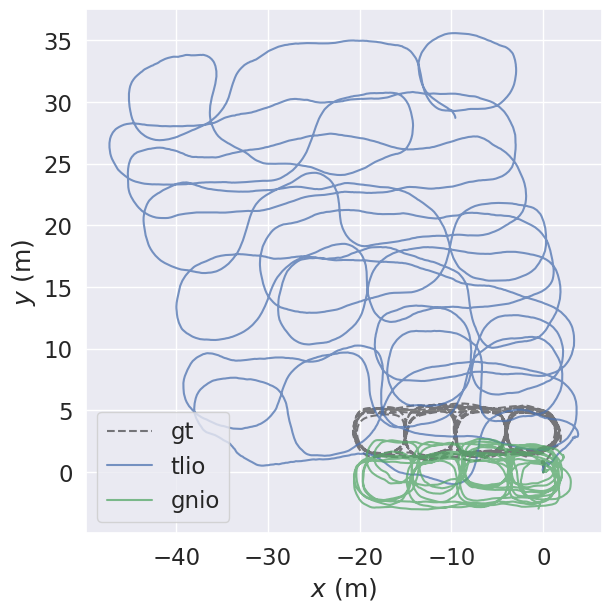}
\par\end{centering}
}%
\end{minipage}\hfill{}%
\begin{minipage}[t]{0.23\linewidth}%
\subfloat[\label{fig:IDOL-B3_UNKNOWN_2}IDOL-B3\_UNKNOWN\_2]{\begin{centering}
\includegraphics[width=1\linewidth,totalheight=5cm,keepaspectratio]{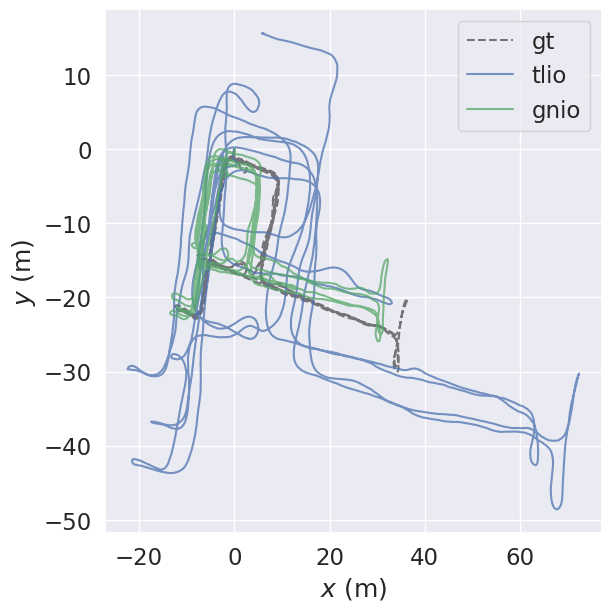}
\par\end{centering}
}%
\end{minipage}

\caption{\label{fig:Qualitative-comparison-of}Qualitative comparison of estimated
trajectories.}
\end{figure*}

\subsubsection{Impact of Motion Bank Size}

We investigate the sensitivity of the GNIO framework to the size of
the Motion Bank ($m$), which determines the number of learnable global
motion prototypes available for the attention mechanism. We varied
$m$ across $\left\{ 16,32,64,128\right\} $ while keeping other hyperparameters
fixed.

As illustrated in \Figref{Impact-of-Motion}, the model performance
exhibits a clear trade-off between expressivity and redundancy:
\begin{itemize}
\item Under-parametrization ($m=16$): With a small bank size, the RMSE
is relatively high ($2.07m$). This suggests that $16$ prototypes
are insufficient to capture the full diversity of human motion modes
(e.g., distinguishing between varying walking speeds, turning, and
stair climbing), leading to mode collapse where distinct activities
are mapped to the same prototype.
\item Optimal Capacity ($m=64$): The error decreases significantly as the
bank size increases, reaching a minimum RMSE of $1.98m$ at $m=64$.
At this capacity, the network learns a sufficiently granular dictionary
of motion patterns to resolve ambiguities in the local sensor data.
\item Over-parametrization ($m=128$): Further increasing the bank size
leads to performance degradation ($2.07m$). A distinct diminishing
returns effect is observed, likely due to prototype redundancy. When
the bank is too large, the attention mechanism suffers from signal
dilution, where the query matches multiple similar prototypes with
low confidence, introducing noise into the context feature $\mathbf{c}_{k}$.
\end{itemize}

\subsection{Qualitative Evaluation}

To provide a qualitative assessment of the trajectory estimation,
we visualize the estimated paths of GNIO and the baseline TLIO \cite{liu2020tlio}
against the ground truth (GT) across several representative sequences
from the four benchmarks. As illustrated in \Figref{Qualitative-comparison-of},
GNIO (shown in green) consistently tracks the ground truth more closely
than the baseline (shown in blue), which suffers from noticeable cumulative
drift.

\subsubsection{Drift Suppression in Pedestrian Motion (\Figref{OxIOD-SLOW_WALKING,OxIOD-HANDBAG})}

In the OxIOD sequences, particularly the Slow Walking and Handbag
scenarios, the baseline TLIO tends to deviate from the reference loop.
In contrast, GNIO maintains a tight trajectory that aligns with the
ground truth. This visual evidence supports our claim that the Gated
Prediction Head effectively acts as a soft-ZUPT, preventing the velocity
leakage that typically occurs during slow or periodic human motion.

\subsubsection{Robustness to Placement and Dynamics (\Figref{RIDI-ZHICHENG_HANDHELD2,RIDI-SHALI_BAG1,IDOL-B1_KNOWN_8,IDOL-B3_UNKNOWN_2})}

The RIDI and IDOL plots demonstrate GNIO’s robustness in challenging
conditions. Even when the IMU is carried in a bag (\figref{RIDI-SHALI_BAG1})
or subjected to the sharp maneuvers and abrupt direction changes characteristic
of the IDOL dataset (\figref{IDOL-B1_KNOWN_8}), GNIO accurately captures
the overall shape and orientation of the movement.

\subsubsection{Long-Range Generalization (\figref{RoNIN-A016_1,RoNIN-A052_2})}

In the unconstrained RoNIN sequences, GNIO exhibits smoother trajectory
reconstruction. By leveraging the Motion Bank to query global motion
prototypes, the system successfully disambiguates complex activities,
preventing the erratic jumps or scale errors observed in the baseline
predictions.

Overall, the qualitative results in \Figref{Qualitative-comparison-of}
confirm that our architectural innovations---specifically the decomposition
of displacement and the use of global semantic context---lead to
statistically consistent and physically plausible trajectory estimates
across diverse real-world scenarios.

\section{Limitations and Feature Work}

While GNIO demonstrates robust performance on pedestrian datasets,
its accuracy degrades significantly when applied to vehicle-based
datasets such as KITTI \cite{geiger2012we}, GEODE\cite{chen2024heterogeneous}.
The primary reason for this degradation lies in the physics of error
propagation at high velocities. Inertial odometry essentially performs
integration of velocity estimates to obtain position. In pedestrian
navigation, typical speeds are low ($\sim1.4m/s$), so a small relative
error in velocity prediction (\textit{e.g.}, $1\%$) results in a
manageable position drift. However, in automotive scenarios where
speeds frequently exceed $10m/s$, the same relative error magnitude
translates into massive absolute position drift. Furthermore, unlike
pedestrian navigation, where frequent ZUPT help reset drift, continuous
high-speed driving lacks such constraints, causing small biases in
the predicted velocity to accumulate rapidly into large position errors.

To address the challenges of high-speed tracking, we propose three
specific directions:

\subsubsection{Long-Sequence Modeling}

We will replace the sliding-window backbone with State Space Models
(\textit{e.g.}, Mamba \cite{gu2024mamba}) to capture long-term acceleration
dependencies, addressing the observability issue during constant-speed
cruising.

\subsubsection{Mixture of Experts}

To handle the drastic dynamic range between pedestrians and vehicles,
we will implement an Mixture of Experts (MoE) architecture with specialized
Low-Speed and High-Speed experts, gated by a dynamic router that identifies
motion context via vibration analysis.

\subsubsection{Kinematic Constraints}

We plan to incorporate non-holonomic constraints into the loss function,
explicitly penalizing velocity estimates to reflect the physical limitations,
such as Ackerman steering geometry.

\section{Conclusion}

In this work, we introduced GNIO, a novel architecture that cures
the curse of drift in inertial navigation through two key structural
innovations. First, the learnable Motion Bank provides global semantic
context by querying a dictionary of motion prototypes, allowing the
network to disambiguate complex activities that local windows cannot
resolve. Second, the Gated Prediction Head decomposes displacement
into magnitude and direction, acting as a differentiable soft-ZUPT
mechanism that proactively suppresses noise during stationarity and
scales predictions during dynamic motion.

Extensive experiments across four major benchmarks demonstrate that
GNIO significantly outperforms state-of-the-art CNN and Transformer-based
models, achieving up to a $60.21\%$ reduction in trajectory error.
By embedding structural motion priors and semantic memory directly
into the network, GNIO provides a more robust and generalized solution
for precise, long-term stable inertial odometry.

\bibliographystyle{IEEEtran}
\bibliography{ref}

\end{document}